%% file: colm2024_conference.tex
\documentclass[dvipsnames]{article} %
\usepackage{colm2024_conference}

\usepackage{booktabs}
\usepackage{enumitem}
\usepackage{wrapfig}
\usepackage{algorithm}
\usepackage{algpseudocode}
\usepackage{graphicx}
\usepackage{subfigure}
\usepackage[misc]{ifsym}

\usepackage{microtype}
\usepackage{amsmath}
\usepackage{colortbl}
\usepackage[utf8]{inputenc}
\usepackage[T1]{fontenc}
\definecolor{lightgray}{rgb}{0.9,0.9,0.9}
\usepackage{caption}
\usepackage{subcaption}
\usepackage{setspace}
\usepackage{url}
\usepackage{multirow}
\usepackage{tabularx}
\usepackage{blindtext}
\usepackage{pgfplots}
\pgfplotsset{compat=1.18} 
\usepackage{tikz}
\usetikzlibrary{er,positioning,bayesnet}
\usepackage{makecell}
\usepackage{tipa}
\usepackage{siunitx}
\usepackage{nicefrac}
\usepackage{listings}
\usepackage[raster,skins, most]{tcolorbox} %
\usepackage{xltabular}
\usepackage{adjustbox}
\usepackage{xurl}
\usepackage{rotating}
\usepackage[normalem]{ulem}

\usepackage{fontawesome}

\useunder{\uline}{\ul}{}

\input{math_commands.tex}

\renewcommand{\ghlink}{https://github.com/Alibaba-NLP/DeepResearch}

\usepackage{makecell}
\usetikzlibrary{tikzmark}
\makeatletter
\newcommand*\myfontsize{%
  \@setfontsize\myfontsize{7}{8}%
}
\makeatother

\usepackage{fontawesome}   % Font Awesome 4.7 macros, includes \faLeaf

\definecolor{uclablue}{RGB}{159, 195, 224}

\definecolor{uclagold}{RGB}{255, 240, 180}

\definecolor{aliceblue}{RGB}{255, 238, 241}

\definecolor{cadmiumgreen}{rgb}{0.0, 0.42, 0.24}

\definecolor{myred}{rgb}{0.7, 0.3, 0.0}
\definecolor{myblue}{rgb}{0.2, 0.3, 0.6}
\definecolor{babygreen}{rgb}{0.85, 0.97, 0.85}

\definecolor{purple1}{RGB}{126, 107, 196}
\definecolor{purple2}{RGB}{199, 158, 207}
\definecolor{purple3}{RGB}{214, 200, 255}
\definecolor{purple4}{RGB}{254, 240, 255}

\definecolor{deepblue}{RGB}{48, 58, 82}

\makeatletter

\newcommand{\symboletongyi}{\raisebox{0pt}{~\includegraphics[scale=0.012]{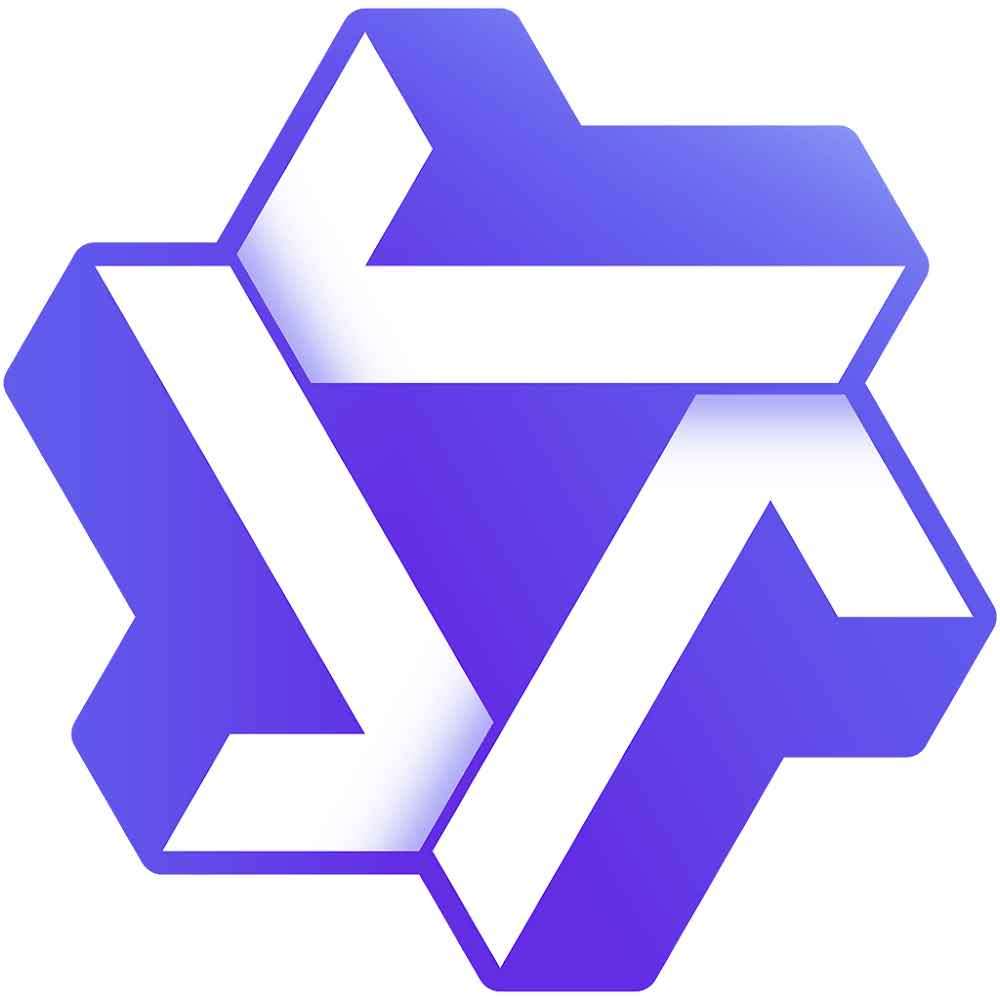}}~}

\definecolor{deepPurple}{HTML}{330066}

\definecolor{uclablue_old}{rgb}{0.15, 0.45, 0.68}
\hypersetup{
    breaklinks,
    citecolor=uclablue_old,
    colorlinks=true,
}

\newtcolorbox{mybox}[2][]
  {colback = black!5!white, colframe = black!75!black, fonttitle = \bfseries,
    colbacktitle = black!100!black, enhanced, before upper={\fontsize{8}{11}\obeyspaces\obeylines\selectfont}, fontupper=\selectfont,
    attach boxed title to top left={yshift=-2.2mm,xshift=4mm},
    title=#2,#1}

\newcommand{\equal}{\textsuperscript{\dag}}

\definecolor{darkgreen}{rgb}{0.0, 0.5, 0.0}

\author{%
\small{Baixuan Li\equal$^{(\textrm{\Letter})}$, Dingchu Zhang\equal, Jialong Wu\equal, Wenbiao Yin$^{(\textrm{\Letter})}$, Zhengwei Tao, Yida Zhao, \\Liwen Zhang, Haiyang Shen, Runnan Fang, Pengjun Xie, Jingren Zhou, Yong Jiang$^{(\textrm{\Letter})}$}%
  \\[1em]               % ← 在上一行末尾插入 1 em 的竖直间距，相当于“空一行”
  {\fontsize{10pt}{11pt}\selectfont          % \large 放大字号；\bfseries 视需要加粗
Tongyi Lab\symboletongyi, Alibaba Group}\\
}

\usepackage{caption}
\usepackage{subcaption}
\usepackage{amssymb}

\definecolor{mydarkgray}{gray}{0.2} % 0=黑, 1=白

\begin{document}

\title{%
\raisebox{-2.17em}{
  \parbox[t]{0.2in}{\includegraphics[width=0.65in]{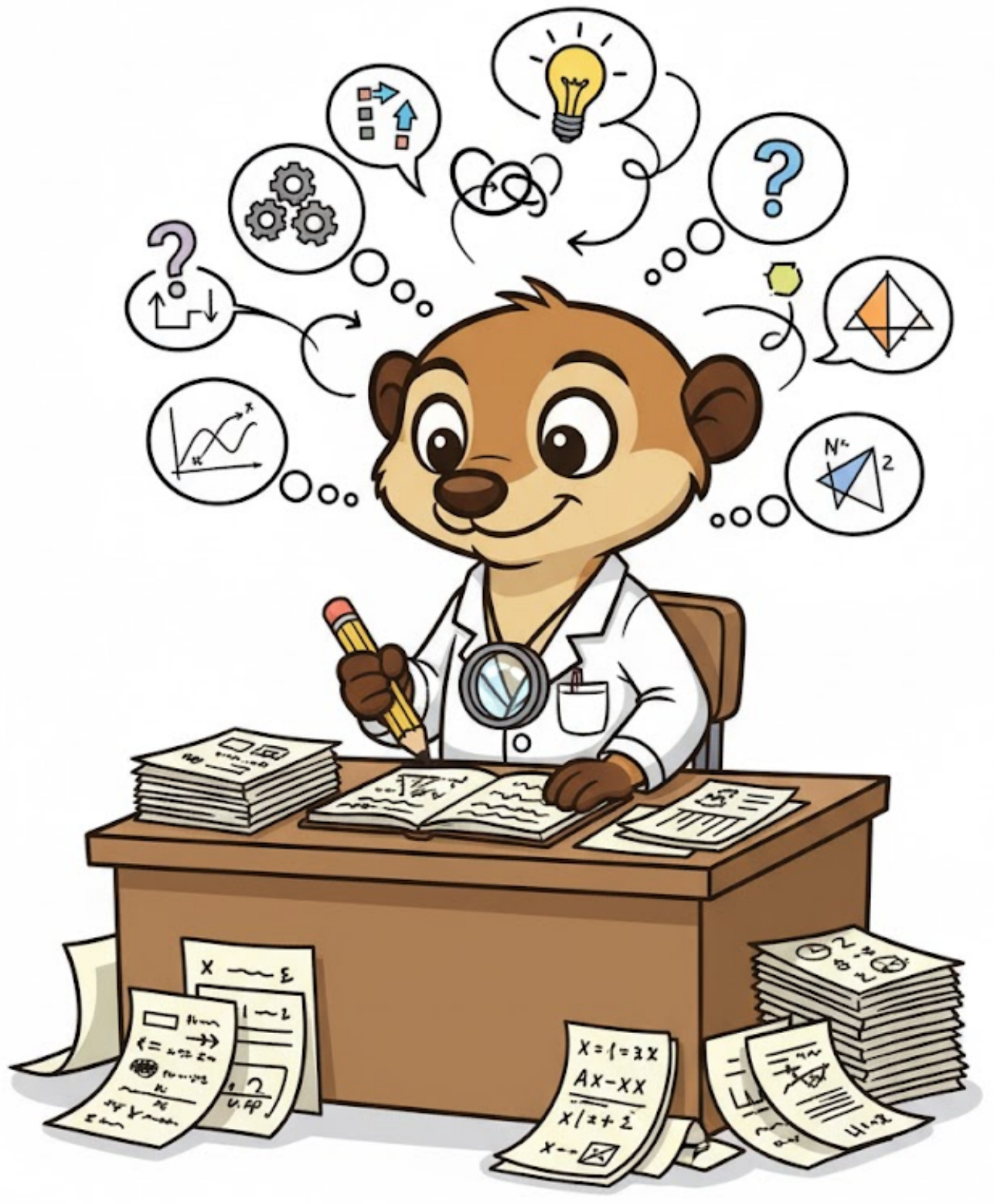}} % 使用 \parbox 放置 logo，并添加一些水平间距
}
\begin{tabular}[t]{l} % 注意这里是 c，不是 l
  \parbox[t]{0.8\textwidth}{\centering % 加了\centering实现居中
    \textsc{ParallelMuse}: Agentic Parallel Thinking\\for Deep Information Seeking
  }
\end{tabular}
}

\maketitle

\begingroup
\renewcommand\thefootnote{\equal}% 去掉编号
\footnotetext{Equal contribution.}
\endgroup

\begingroup
\renewcommand\thefootnote{$^{\textrm{\Letter}}$}% 去掉编号
\footnotetext{Correspondence to: \texttt{baixuan@seu.edu.cn}, \texttt{\{yinwenbiao.ywb, yongjiang.jy\}@alibaba-inc.com}.}
\endgroup

\begin{abstract}
  \input{sections/0_abstract}
\end{abstract}

\input{sections/1_introduction}
\input{sections/2_preliminary}
\input{sections/3_method}
\input{sections/4_experiments}

\input{sections/5_related_work}

\input{sections/6_conclusion}

\input{sections/7_limitation}

\clearpage

\clearpage
\bibliography{custom}
\bibliographystyle{colm2024_conference}

\clearpage
\appendix
% \section{Discussions}
% \label{app:discuss}
% \paragraph{Eval}

\end{document}

%% file: math_commands.tex
\usepackage{amsmath,amsfonts,bm}

\def\eqref#1{equation~\ref{#1}}
\def\1{\bm{1}}

\DeclareMathAlphabet{\mathsfit}{\encodingdefault}{\sfdefault}{m}{sl}
\SetMathAlphabet{\mathsfit}{bold}{\encodingdefault}{\sfdefault}{bx}{n}

%% file: sections/0_abstract.tex
Parallel thinking expands exploration breadth, complementing the deep exploration of information-seeking (IS) agents to further enhance problem-solving capability. However, conventional parallel thinking faces two key challenges in this setting: inefficiency from repeatedly rolling out from scratch, and difficulty in integrating long-horizon reasoning trajectories during answer generation, as limited context capacity prevents full consideration of the reasoning process.
To address these issues, we propose \textbf{\textsc{ParallelMuse}}, a two-stage paradigm designed for deep IS agents. The first stage, \textit{\textbf{Functionality-Specified Partial Rollout}}, partitions generated sequences into functional regions and performs uncertainty-guided path reuse and branching to enhance exploration efficiency. The second stage, \textit{\textbf{Compressed Reasoning Aggregation}}, exploits reasoning redundancy to losslessly compress information relevant to answer derivation and synthesize a coherent final answer. Experiments across multiple open-source agents and benchmarks demonstrate up to \textbf{62\%} performance improvement with a \textbf{10--30\%} reduction in exploratory token consumption.

%% file: sections/1_introduction.tex
\section{Introduction}
Deep information-seeking (IS) agents\footnote{The agents discussed in this work are function-calling agents that adhere to the standard ReAct~\citep{yao2023react} paradigm, operating through an iterative think $\rightarrow$ tool call loop.}~\citep{openaidr,kimi_researcher,tongyidr} can actively uncover hard-to-access information, extending large language models beyond static training data and empowering them to reason over real-world knowledge. This capability emerges from a continual loop of environmental\footnote{This work focuses on deep information-seeking agents, where the term ``environment'' specifically refers to the web environment or information sources with which the agent interacts.} interaction and internal reasoning, through which the agent incrementally builds reasoning depth within a single execution to effectively solve complex problems~\citep{webwalker,wu2025webdancerautonomousinformationseeking,li2025websailornavigatingsuperhumanreasoning,tao2025webshaper,li2025websailorv2bridgingchasmproprietary}. 
In this setting, parallel thinking provides a natural form of test-time scaling: by expanding the number of parallel exploration paths, it broadens the agent’s search while maintaining reasoning depth along each path, thereby enhancing overall performance without altering model parameters.

As commonly recognized, parallel thinking can be viewed as a two-stage process~\citep{li2025raspberry}, involving an initial stage of exploratory sampling and a subsequent stage dedicated to answer generation across sampled candidates.
In this work, we extend this paradigm to the setting of deep IS agents, referred to as \textbf{\textsc{ParallelMuse}}.
We analyze how the characteristics of each stage manifest under agentic conditions and propose systematic optimization strategies derived from these pilot observations.

\textbf{First}, in the \underline{exploratory sampling stage}, conventional rollout strategies in parallel thinking typically restart from scratch at each iteration, resampling the entire exploration space~\citep{skywork2025deepresearch,fu2025deep,zeng2025pushing}. During certain reasoning phases, however, exploration diversity is inherently low, making repeated rollouts inefficient and token-expensive. Prior work introduces \textit{partial rollout} methods that estimate exploration potential via uncertainty and selectively branch where uncertainty is high~\citep{hou2025treerl,arpo,li2025treepo}, but these approaches assume functional homogeneity across tokens, implying that all tokens contribute equally to exploration and exhibit similar uncertainty.

This assumption holds in purely reasoning-oriented tasks such as mathematics or coding but fails in agentic IS settings, where the model must generate both reasoning and tool-call actions. These behaviors naturally form distinct \textit{functional regions} with different uncertainty patterns.
Motivated by this observation, we propose the \textit{\textbf{Functionality-Specified Partial Rollout}} method as the first stage of the \textsc{ParallelMuse} framework. 
The method segments the generated sequence into functional regions, estimates uncertainty independently within each, and selectively expands rollouts for reasoning steps with higher exploration potential. 
This enables behavior-level estimation of exploration potential, allowing targeted exploration across different functional behaviors, and improving overall efficiency in agentic tasks.

\textbf{Second}, in the \underline{answer generation stage}, parallel thinking produces multiple reasoning candidates from which a single answer must be derived, typically through \textit{answer selection}~\citep{self_consistency,fu2025deep} or \textit{answer aggregation}~\citep{jiang2023llm,liang2024encouraging,zhang2025and,qiao2025webresearcher}. 
In complex agentic tasks with vast sampling spaces, the correct answer may not dominate numerically, as it often constitutes only a small fraction of all possible sampled outcomes. 
Moreover, the continual incorporation of external, non–model-generated information shifts the output distribution, further hindering confidence calibration~\citep{jang2024calibrated}.
As a result, majority voting~\citep{self_consistency} and confidence-based selection~\citep{fu2025deep} often fail.
For answer aggregation, focusing only on final answers neglects intermediate reasoning, while incorporating entire reasoning traces is infeasible for long-horizon agents due to context limits. 
Recent work~\citep{qiao2025webresearcher} seeks a compromise by aggregating only the last few reasoning steps, but this discards earlier content, which often reflects planning and problem decomposition and is essential for evaluating the coherence of the final answer.

To address this challenge, we first conceptualize the IS task as a process of discovering key entities and building connections among them~\citep{li2025websailornavigatingsuperhumanreasoning, tao2025webshaper, li2025websailorv2bridgingchasmproprietary}. Based on our preliminary observations, only a small portion of the entities explored by an agent contribute meaningfully to the final answer, revealing substantial redundancy and strong potential for lossless compression in the generated interaction–reasoning trajectories. Building on this insight, we propose the \textbf{\textit{Compressed Reasoning Aggregation}} method as the second stage of the \textsc{ParallelMuse} paradigm. 
The method first condenses all reasoning candidates into concise, structured reports that preserve only information relevant to answer derivation, and then aggregates these compressed reports to produce the final answer. This approach enhances processing efficiency and mitigates the bias of majority-based selection, enabling more reliable and coherent answer generation.

The proposed \textsc{ParallelMuse} is evaluated on four open-source deep IS agents, including GPT-OSS-20B~\citep{gptoss}, GPT-OSS-120B, DeepSeek-V3.1-Terminus~\citep{deepseekv3}, and Tongyi-DeepResearch-30B-A3B~\citep{tongyidr}, across four challenging benchmarks that jointly assess deep search and reasoning abilities: BrowseComp~\citep{bc_en}, BrowseComp-zh~\citep{bc_zh}, GAIA~\citep{mialon2023gaia}, and  Humanity's Last Exam (HLE)~\citep{hle}. Extensive experiments show that \textsc{ParallelMuse} achieves up to \textbf{62\%} improvement while requiring only \textbf{70--90\%} of the exploratory token cost of conventional parallel thinking. Beyond the empirical gains, our analysis provides key insights into the mechanisms of deep IS agents, offering guidance for future research in agentic reasoning.

%% file: sections/2_preliminary.tex
\section{Pilot Observation}

We begin with a preliminary analysis of the characteristics of deep information-seeking (IS) agents and their associated tasks, providing insights from two perspectives: (i) exploratory sampling (trajectory rollout) and (ii) the resulting interaction–reasoning trajectories.

\subsection{Distinct Uncertainty Patterns Across Functional Reasoning Steps}\label{subsec:uncertainty-pattern}

In deep IS tasks, agents must not only reason over internal knowledge but also explore unknown information through tool use and environmental interaction. 
% This distinguishes them from conventional reasoning models that operate solely on internal knowledge.
While pure reasoning models use tokens exclusively for internal reasoning, deep IS agents additionally allocate tokens for tool invocation to retrieve external information, reflecting distinct functional roles in token utilization.

Formally, each step consists of a reasoning segment, a tool invocation, and its corresponding tool response. We denote the set of tokens generated by the model at step $t$ as $\mathcal{T}_t=\{x_{t,1},x_{t,2},...,x_{t,m}\}$, with $x_{t,i}$ denoting the $i$-th token.
The set is partitioned into two subsets: $\mathcal{T}^{r}_t$, representing \textit{reasoning} tokens, and $\mathcal{T}^{e}_t$, representing \textit{exploration} tokens. 
This partition holds for each step, implying $\mathcal{T}^{r}_t \cup \mathcal{T}^{e}_t=\mathcal{T}_t$.
% and $\mathcal{T}^{r}_t \cap \mathcal{T}^{e}_t=\emptyset$.
By extension, aggregating these sets across the entire trajectory yields global sets ($\mathcal{T}, \mathcal{T}^r, \mathcal{T}^e$).
In contrast, a pure reasoning task would have an empty exploration set, $\mathcal{T}^{e}=\emptyset$, satisfying $\mathcal{T} = \mathcal{T}^r$.

Furthermore, we observe that the uncertainty associated with tokens in the $\mathcal{T}^{r}$ and $\mathcal{T}^{e}$ subsets exhibits distinct temporal dynamics during the agentic interaction-reasoning process.
To quantitatively capture this behavior, we use the \textbf{perplexity} (PPL) of each reasoning step, which is defined as the average PPL of tokens within step $t$, as a proxy for the deep IS agent’s self-uncertainty. 
\begin{equation}\label{ppl_computation}
\text{PPL}(f,t) = \exp\left(-\frac{1}{|\mathcal{T}^{f}_t|}\sum_{i=0}^{|\mathcal{T}_t|}\log p\ \left(x_{t,i}\mid x_{<t,i}\right)\right), \quad x_{t,i}\in |\mathcal{T}^f_t|,
\quad f\in\{r,e\},
\end{equation}

where $f$ represents the functional region of the entire trajectory, which is partitioned into a reasoning region $r$ and an exploration region $e$.

\begin{figure}[h]
    \centering
    \includegraphics[width=\textwidth]{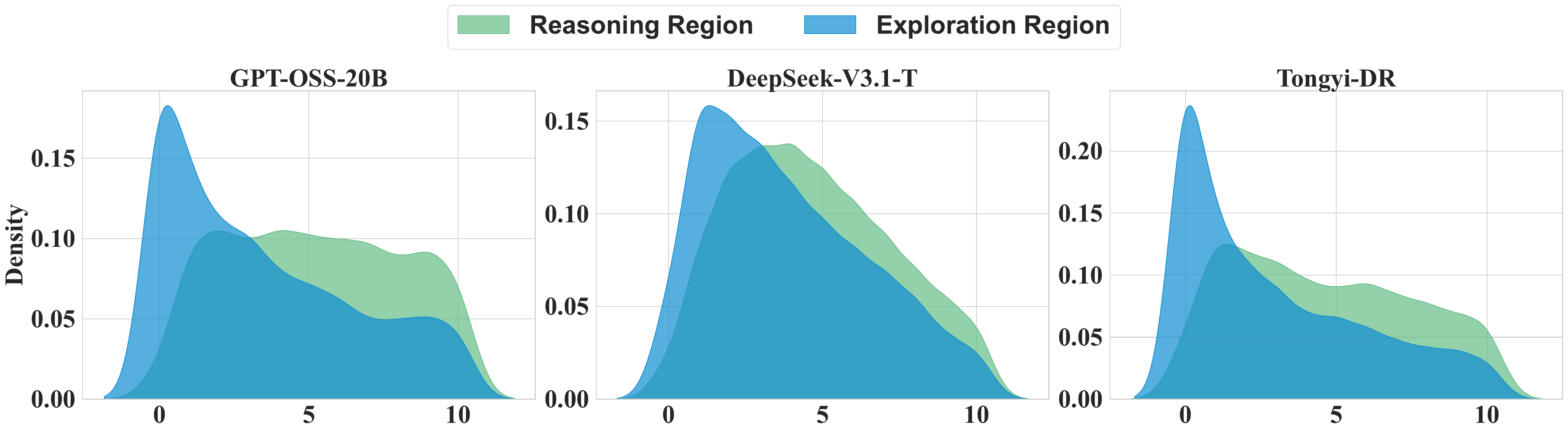}
    \caption{KDE-smoothed distribution of steps with top-4 uncertainty on the BrowseComp subset (truncated to earlier steps as later ones are typically more certain). DeepSeek-V3.1-T denotes DeepSeek-V3.1-Terminus, and Tongyi-DR denotes Tongyi-DeepResearch-30B-A3B.}
    \label{fig:think_tool_density}
\end{figure}

We analyze $\text{PPL}(r,t)$ and $\text{PPL}(e,t)$ across steps to characterize the distinct uncertainty dynamics of reasoning and exploration within the agentic reasoning-interaction process. 
As shown in Figure~\ref{fig:think_tool_density}, across multiple deep IS models, we examine the distribution of the top-4 uncertainty steps observed during task execution. 
The results reveal a consistent pattern: exploration uncertainty reaches its highest levels at the earliest stages, when no external information has yet been gathered, while reasoning uncertainty peaks slightly later as the agent begins integrating retrieved information into its internal reasoning process.

Specifically, \textit{exploration uncertainty} tends to peak at the beginning of the task, when the agent has not yet gathered external information and must explore the environment under minimal prior knowledge. 
\textit{Reasoning uncertainty}, in contrast, reaches its peak slightly later—still in the early stage—when the agent begins to process and integrate the newly retrieved information into its internal reasoning chain. As the task proceeds, both forms of uncertainty gradually decline as the agent accumulates knowledge and its reasoning process becomes more grounded, resulting in increasingly confident decisions and actions.

This observation further informs the design of the \textbf{\textit{Functionality-Specified Partial Rollout}} method in \textsc{ParallelMuse}, which enhances agentic parallel thinking by enabling more efficient exploration.

\subsection{From Exploration Redundancy to Losslessly Compressible Trajectory}\label{sec:redundancy}

Following recent studies~\citep{li2025websailornavigatingsuperhumanreasoning,tao2025webshaper,li2025websailorv2bridgingchasmproprietary}, deep IS tasks can be formulated as a process of \textit{entity discovery} and \textit{relation construction}. 
Formally, given an initial query or objective $q$, the agent incrementally builds a set of discovered entities 
$\mathcal{V} = \{v_1, v_2, \ldots, v_N\}$ through iterative interactions with external information sources. 
At each step $t$, the agent performs exploration to retrieve candidate entities 
$\tilde{\mathcal{V}}_t$ and reasoning to determine their relevance and relational connections. 
The evolving information state of the agent can thus be expressed as:
\begin{equation}\label{graph}
\mathcal{G}_t = (\mathcal{V}_t, \mathcal{R}_t),
\end{equation}
where $\mathcal{V}_t$ denotes the set of \textit{effective entities} (i.e., entities considered valid for answer derivation) and 
$\mathcal{R}_t \subseteq \mathcal{V}_t \times \mathcal{V}_t$ represents the relations constructed among them. 
The goal of the task is to iteratively refine $\mathcal{G}_t$ until it contains the key entities and connections 
necessary for deriving the final answer. 
Once the entire agentic reasoning process terminates, the final graph 
$\mathcal{G}_{\text{final}} \supseteq \mathcal{I}_{\text{answer}}$ 
(where $\mathcal{I}_{\text{answer}}$ denotes all information essential for answer derivation), 
serving as the core representation of the reasoning trajectory.

Based on this formulation of deep IS tasks, we approximate the redundancy of a task’s complete reasoning trajectory by measuring the proportion of \textit{effective entities}—those that contribute directly or indirectly to answer derivation—within all entities discovered during execution. 
Formally, let $\mathcal{V}_{\text{total}}$ denote the set of all entities explored by the agent during a task, and let $\mathcal{V}_{\text{eff}} \subseteq \mathcal{V}_{\text{total}}$ represent the subset of entities that are directly or indirectly useful for deriving the final answer. 
The redundancy ratio $\Gamma_{\text{red}}$ is defined as:
\begin{equation}
\Gamma_{\text{red}} = 1 - \frac{|\mathcal{V}_{\text{eff}}|}{|\mathcal{V}_{\text{total}}|}.
\end{equation}
A higher $\Gamma_{\text{red}}$ indicates greater redundancy in the agent’s interaction–reasoning trajectory, which can be interpreted as a stronger potential for \textit{lossless compression} of the reasoning process. 
In this context, lossless compression refers to reducing redundant entities and reasoning steps while preserving all 
\begin{wrapfigure}{r}{0.6\linewidth}
% \vspace{-4em} % 上方间距
    \centering
    \includegraphics[width=0.99\linewidth]{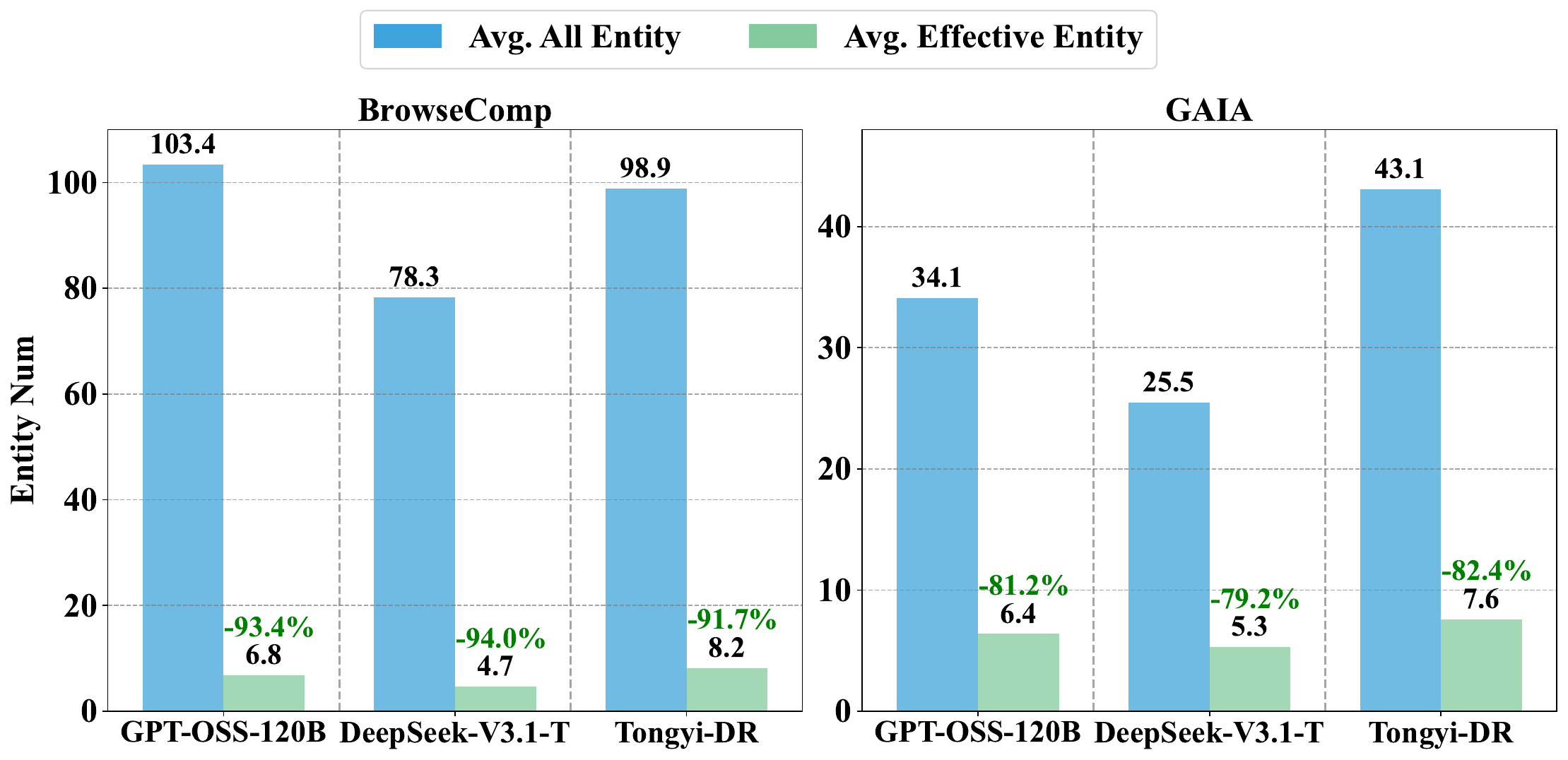}
    % \vspace{-2em}
    \caption{Average entity count per task and per model, where entities are extracted by GPT-4.1 based on the complete reasoning trajectory and ground-truth answer.}
    % \vspace{-2em} % 下方间距
    \label{fig:entity_efficiency}
\end{wrapfigure}
reasoning information necessary for the complete derivation of the final answer (i.e., extracting and representing $\mathcal{G}_{\text{final}}$), and thus $\Gamma_{\text{red}}$ can be regarded as an approximate indicator of the degree of lossless compressibility.

Accordingly, we compute the reasoning trajectory redundancy of several mainstream deep IS agents during real task execution. As illustrated in Figure \ref{fig:entity_efficiency}, all models exhibit consistently high redundancy, indicating that the reasoning trajectories in deep IS tasks are highly losslessly compressible. This observation supports the design of the \textbf{\textit{Compressed Reasoning Aggregation}} method in \textsc{ParallelMuse}, which aims to  integrate as much effective reasoning information as possible into final aggregation with minimal information loss.

%% file: sections/3_method.tex
\begin{figure}[h]
    \centering
    \includegraphics[width=\textwidth]{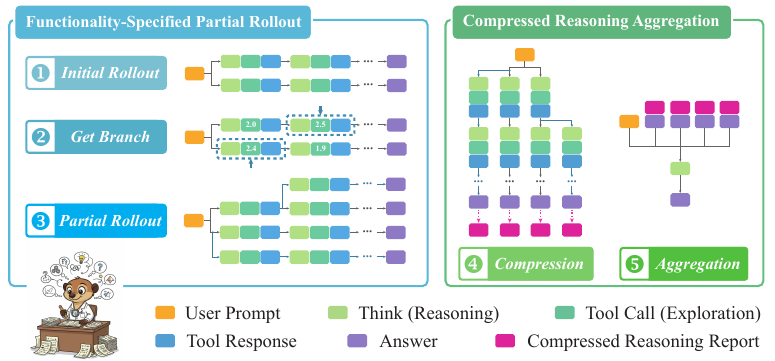}
    \caption{Workflow of \textsc{ParallelMuse}, including (\textit{Left}) the Functionality-Specified Partial Rollout, where the \textit{Get Branch} shows the selection of top-$k$ steps based on (exploration) tool-call uncertainty (just as an example of branching criterion), and (\textit{Right}) the Compressed Reasoning Aggregation.}
    \label{fig:method}
\end{figure}

\section{ParallelMuse}

The proposed \textsc{ParallelMuse} is a two-stage agentic parallel thinking paradigm comprising two complementary components: (i) \textit{\textbf{Functionality-Specified Partial Rollout}} and (ii) \textit{\textbf{Compressed Reasoning Aggregation}}. As shown in Figure \ref{fig:method}, these correspond to the first stage of exploratory sampling and the second stage of answer generation in the overall parallel thinking process, respectively.

\subsection{Functionality-Specified Partial Rollout}

\paragraph{Functionality-Specified Branching Step Identification.}

Agent models inherently partition generated tokens into functional regions, typically reasoning and exploration, signaled by special tokens such as \texttt{<think>} and \texttt{<tool\_call>}.
We leverage these markers to identify distinct functional segments within the generation process.
To enable more targeted partial rollout, it is essential to identify reasoning steps with higher exploration potential. We measure this potential through the model’s generation uncertainty at each step, as higher uncertainty indicates greater diversity in possible continuations and thus a broader exploration space. Accordingly, we compute the \textbf{step-level perplexity} (PPL) within each functional region, as defined in \eqref{ppl_computation}, to quantify the agent’s generation uncertainty. 

This process is conducted in an offline manner to ensure optimal branch selection. As shown in Figure \ref{fig:method} (\textit{Left}), we first generate $M$ initial trajectories from scratch, compute step-level PPL for reasoning and exploration regions along these trajectories for each step, and select the top-$k$ steps with the highest uncertainty in the chosen functional region $f \in \{r,e\}$ (defined in \eqref{ppl_computation}) as branching points for subsequent partial rollouts. It is noted that $M$, $k$, and $f$ are tunable hyper-parameters.

\paragraph{Asynchronous Partial Rollout.}

From the selected high-uncertainty steps, \(N - M\) additional partial rollouts are launched asynchronously to expand exploration, where \(N\) is the overall sampling budget. 
Each branch directly reuses the preceding context rather than regenerating it from scratch, continuing from cached hidden states in the key–value (KV) cache~\citep{li2024survey,wu-etal-2025-scope}. 
This reuse eliminates redundant forward passes, yielding substantial savings in both token and compute cost. 

We implement an asynchronous rollout engine to parallelize branch generation while preserving each branch’s causal decoding consistency, enabling multiple branches to expand concurrently and efficiently. 

The acceleration comes from two sources: (i) \emph{prefix reuse} via KV caching and (ii) \emph{asynchronous parallelization}. 
Let branch $j$ reuse a prefix of token length $p_j$ and generate a suffix of token length $s_j$. 
With cold decoding (no KV reuse), the cost is $C^{\text{cold}}_j = c_{\text{cold}}\cdot(p_j + s_j)$; 
with KV reuse, the cost is $C^{\text{hot}}_j = c \cdot s_j$.
Here, $c>0$ denotes the per-token compute cost under cached decoding (with KV reuse), and $c_{\text{cold}}\ge c$ denotes the per-token cost when regenerating from scratch (without KV reuse).
\begin{equation}
\label{eq:reuse-factor}
\text{ReuseFactor} \equiv
\frac{\sum_{j} C^{\text{cold}}_j}{\sum_{j} C^{\text{hot}}_j}=\frac{c_{\text{cold}}}{c}\left(1 +\frac{\sum_{j} p_j}{\sum_{j} s_j}\right).
\end{equation}
Asynchronous scheduling parallelizes hot decoding across $P$ active branches. 
If $\alpha \in [0,1]$ denotes the parallelizable ratio, the throughput gain obeys the Amdahl-type bound~\citep{amdahl1967validity}:
\begin{equation}
\label{eq:para-factor}
\text{ParaFactor}(P)\le\frac{1}{(1-\alpha) + \alpha/P}.
\end{equation}
Combining \eqref{eq:reuse-factor} and \eqref{eq:para-factor} yields the overall speedup: $\text{Speedup}_{\text{total}}\lesssim \text{ReuseFactor}\times \text{ParaFactor}(P)$.
In the practical regime with efficient KV caching ($c \approx c_{\text{cold}}$), high parallelizability ($\alpha \approx 1$), and $P$ within hardware concurrency, this simplifies to
\begin{equation}
\text{Speedup}_{\text{total}}\approx\left(1 + \frac{\sum_{j} p_j}{\sum_{j} s_j}\right) P.
\end{equation}

This design jointly exploits deterministic prefix reuse and asynchronous parallelization to achieve near-linear speedup in exploration efficiency with relatively lower token cost.

\subsection{Compressed Reasoning Aggregation}

\paragraph{Structured Report-Style Compression.}

Building on the observations in Section \ref{sec:redundancy}, we note that the complete reasoning trajectories obtained after the first-stage exploratory sampling in deep information-seeking (IS) tasks exhibit high redundancy, implying strong lossless compressibility with respect to answer derivation. To effectively integrate richer intermediate reasoning information during the answer aggregation stage while maintaining computational efficiency and avoiding context overflow, we first compress each candidate reasoning trajectory produced in the exploratory stage.

As shown in Figure \ref{fig:method} (\textit{Right}), for each reasoning trajectory generated in solving deep IS tasks, the compression objective is to produce a structured report that preserves key elements essential to answer derivation. The report records: 

\begin{enumerate}[label=\textbf{(\roman*)}]
    \item \textbf{Solution Planning:} describes how the main problem is decomposed into subproblems, 
    including their dependency structure and execution order.

    \item \textbf{Solution Methods:} specifies the tools invoked to solve each subproblem, 
    the corresponding parameters used, and any subanswers that contribute directly or indirectly to the final solution.

    \item \textbf{Final Reasoning:} illustrates how the identified subproblems and associated subanswers 
    are integrated to derive the final answer.
\end{enumerate}

Irrelevant exploratory content, including redundant tool responses and ineffective reasoning or tool calls, is removed. This process effectively reconstructs the agent’s internal information state graph $\mathcal{G}$ (defined in \eqref{graph}), which comprehensively captures all information relevant to answer derivation.

\paragraph{Reasoning-Guided Answer Aggregation.}

After obtaining $N$ compressed reports from the exploratory sampling stage, we can jointly consider all $N$ globally compressed reasoning candidates within the limited context window, rather than focusing only on their final answers or partial reasoning traces. This enables a more comprehensive evaluation of reasoning coherence and supports a more reliable determination of the optimal answer. In this aggregation stage, we explicitly prevent the model from relying solely on answer consistency as a correctness signal, mitigating the bias toward majority answers and ensuring that reasoning coherence remains the primary criterion. Furthermore, we restrict the model from trivially concatenating or enumerating different answers to preserve aggregation validity.

It is also important to note that each report already contains sufficient tool-calling provenance and attribution information for answer derivation. Therefore, during the aggregation phase, the model does not perform additional tool invocations for secondary verification but instead conducts reasoning purely over the information encoded in the $N$ reports. Empirical results later demonstrate that this approach is both effective and computationally efficient.

%% file: sections/4_experiments.tex
\section{Experiments}

We focus on evaluating the effectiveness and efficiency of applying \textsc{ParallelMuse} to existing deep information-seeking (IS) agents.
Comprehensive experiments are conducted to examine the impact of its two stages both individually and jointly, assessing how each contributes to overall task performance.

\subsection{Setup}\label{sec:setup}

\paragraph{Benchmarks.} We evaluate \textsc{ParallelMuse} on four challenging deep IS benchmarks: BrowseComp~\citep{bc_en}, BrowseComp-zh~\citep{bc_zh}, GAIA~\citep{mialon2023gaia}, and Humanity's Last Exam (HLE)~\citep{hle}. These benchmarks jointly assess both deep search and reasoning capabilities, with BrowseComp and BrowseComp-zh placing greater emphasis on deep search, HLE focusing more on reasoning, and GAIA providing a balanced evaluation across both dimensions.

For efficient text-only evaluation, we use sampled subsets from large-scale datasets: 200 randomly selected tasks from BrowseComp, 157 search-focused text-only tasks from HLE, and 103 text-only tasks from GAIA~\citep{Li2025webthinker}, while using the full 289-task set for BrowseComp-zh.

\vspace{-0.5em}

\paragraph{Tools.} 
We adopt the standard tool configuration commonly used in deep IS agents~\citep{wu2025webdancerautonomousinformationseeking, li2025websailornavigatingsuperhumanreasoning,tao2025webshaper,li2025websailorv2bridgingchasmproprietary,qiao2025webresearcher}, which includes two core tools for interacting with the web environment and retrieving external information:
\begin{itemize}
    \vspace{-0.5em}
    \item \textbf{Search:} Performs batched Google queries and returns the top-10 ranked results for each.
    \item \textbf{Visit:} Fetches webpages from multiple URLs and extracts information relevant to the given goal.
\end{itemize}

\vspace{-1em}

\paragraph{Agent Models.} We select four open-source agent models with diverse parameter scales and advanced tool-use capabilities for deep IS tasks: GPT-OSS-20B~\citep{gptoss}, GPT-OSS-120B, DeepSeek-V3.1-Terminus (DeepSeek-V3.1-T, 671B)~\citep{deepseekv3}, and Tongyi-DeepResearch-30B-A3B (Tongyi-DR-30B-A3B)~\citep{tongyidr}. All agent models are invoked under the official function-calling protocol. Unless otherwise specified, we use the \textit{same} agent model to perform both stages of the \textsc{ParallelMuse}.

\vspace{-0.5em}

\paragraph{Baselines.} In addition to the standard inference baseline without any parallel thinking \underline{(\textit{No Scaling})}, we compare \textsc{ParallelMuse} against several mainstream parallel thinking baselines. These include:
\begin{wraptable}{r}{5.5cm}
\small
\centering
\caption{Default settings of our proposed \textsc{ParallelMuse}.}
\renewcommand{\arraystretch}{1.3} % 调整行高
\begin{adjustbox}{width=1.0\linewidth}
\centering
\label{tab:hyperparams}
\begin{tabular}{l|c}
\toprule
\textbf{Hyper-Parameters} & \textbf{Values} \\
\midrule
Sampling Budget $N$ & 8 \\
\#Initial Rollout $M$ & 1 \\
Branching PPL Top-$K$ & 2 \\
\#Branching Times per Step & 3 \\
\bottomrule
\end{tabular}
\end{adjustbox}
\end{wraptable}
(i) \underline{Self-Consistency (\textit{Majority Vote})}~\citep{self_consistency}, which selects the most frequent answer across multiple trajectories as the final output;
(ii) \underline{\textit{Max \#Tool Call}}~\citep{zeng2025pushing}, which heuristically chooses the answer derived from the trajectory with the largest number of environment interactions; and
(iii) \underline{DeepConf (\textit{Weighted Vote})}~\citep{fu2025deep}, which weights answers by the model’s confidence over each trajectory and selects the answer with the highest final score.

\vspace{-0.5em}

\paragraph{Evaluation Metrics and Hyper-Parameters.} All evaluations are performed under the LLM-as-a-Judge paradigm~\citep{llmasajudge}, using the official evaluation prompts and judging models specified by each benchmark’s released configuration. For the \textit{No Scaling} method, we report the average pass rate over $N$ independent rollouts, while for parallel thinking methods, which yield a single final answer from $N$ rollouts, we report the pass rate of that final output.
For our proposed \textsc{ParallelMuse}, unless otherwise specified, the default hyper-parameter settings are listed in Table \ref{tab:hyperparams}.
To ensure fair comparison and reproducibility, all agent model–specific hyper-parameters are aligned with their official optimal configurations for tool usage.

\subsection{Overall Performance}

\input{table/00_main_results}

We report the performance of closed-source deep IS agents across all benchmarks and compare them with open-source agents equipped with our proposed \textsc{ParallelMuse} and several representative parallel thinking baselines. 
As shown in Table~\ref{tab:main_result}, \textsc{ParallelMuse} consistently achieves the highest performance gains over all baselines across every agent model and benchmark. 
Notably, when applied to Tongyi-DR-30B-A3B, it attains performance comparable to or surpassing that of most closed-source agents.

It is important to note that we further observe that the \textit{Weighted Vote}, which relies on self-estimated confidence, underperforms \textit{Majority Vote} across all models except Tongyi-DR-30B-A3B. 
This can be attributed to confidence miscalibration in agentic settings: as agents repeatedly integrate external, non–model-generated content (e.g., tool responses), their internal probability distributions shift, degrading the reliability of confidence scores~\citep{jang2024calibrated,chhikara2025mind}. 
The exception is the HLE benchmark, which emphasizes reasoning with limited external interaction, and Tongyi-DR-30B-A3B, which benefits from continual pre-training~\citep{su2025scaling} that improves calibration over Search and Visit tool responses.

In contrast, our proposed \textsc{ParallelMuse} avoids confidence-based answer selection altogether, mitigating this source of bias and yielding consistent and substantial improvements across all agent models.

\subsection{Analysis of Partial Rollout over Distinct Functional Regions}\label{sec:function_region}

\input{table/01_partial_rollout}

To analyze the effect of identifying branching steps based on different functional regions in partial rollout, we report the average pass rate after 8 rollouts, as shown in Table \ref{tab:partial_rollout}. The \textit{From Scratch} setting denotes full rollouts without reusing context, where functional region selection is not applicable.
For our proposed \textsc{ParallelMuse}\footnote{We omit results of partial rollout without functional-region distinction (i.e., treating all tokens as homogeneous), as our preliminary experiments show that this setting performs comparably to full from-scratch rollouts and provides no observable gains from branching at high-uncertainty steps.}, we evaluate three functionality-specified partial rollout strategies: using uncertainty from the \textit{Reasoning} region only, using uncertainty from the \textit{Exploration} region only, and a \textit{Mixed} configuration where half of the top-2 branching steps are selected from each region. The \textit{Mixed} setting acts as a compromise, integrating uncertainty cues from both regions.

The results show that the effectiveness of branching based on different functional-region uncertainties varies across models, reflecting their inherent behavioral and capability differences. For instance, GPT-OSS-120B benefits less from reasoning-based branching, as its strong adaptive reasoning mechanism already yields consistently high-quality reasoning with limited exploration potential. In contrast, DeepSeek-V3.1-T employs function calling outside of the \texttt{thinking} mode, resulting in weaker internal reasoning capacity and thus higher sampling potential in reasoning steps. These insights heuristically inform our choice of functional-region uncertainty for partial rollout in \textsc{ParallelMuse}.

We also observe that partial rollout consistently outperforms full from-scratch rollout in most cases. This improvement arises from more targeted exploration. In deep IS tasks, where interaction with the web environment induces an extremely large exploration space, unguided rollouts struggle to identify effective search paths and often fall into local optima. In contrast, uncertainty-guided partial rollout functions analogously to Monte-Carlo Tree Search (MCTS)~\citep{browne2012survey}: while MCTS reuses high-reward trajectories, \textsc{ParallelMuse} reuses low-uncertainty (low-potential) paths and selectively expands exploration at high-uncertainty steps. This strategy allocates the limited sampling budget toward regions with greater expected exploration gain, enhancing both efficiency and effectiveness.

\subsection{Performance Gains from Compressed Reasoning Aggregation}

\begin{wrapfigure}{r}{9cm}
    \centering
    \includegraphics[width=0.99\linewidth]{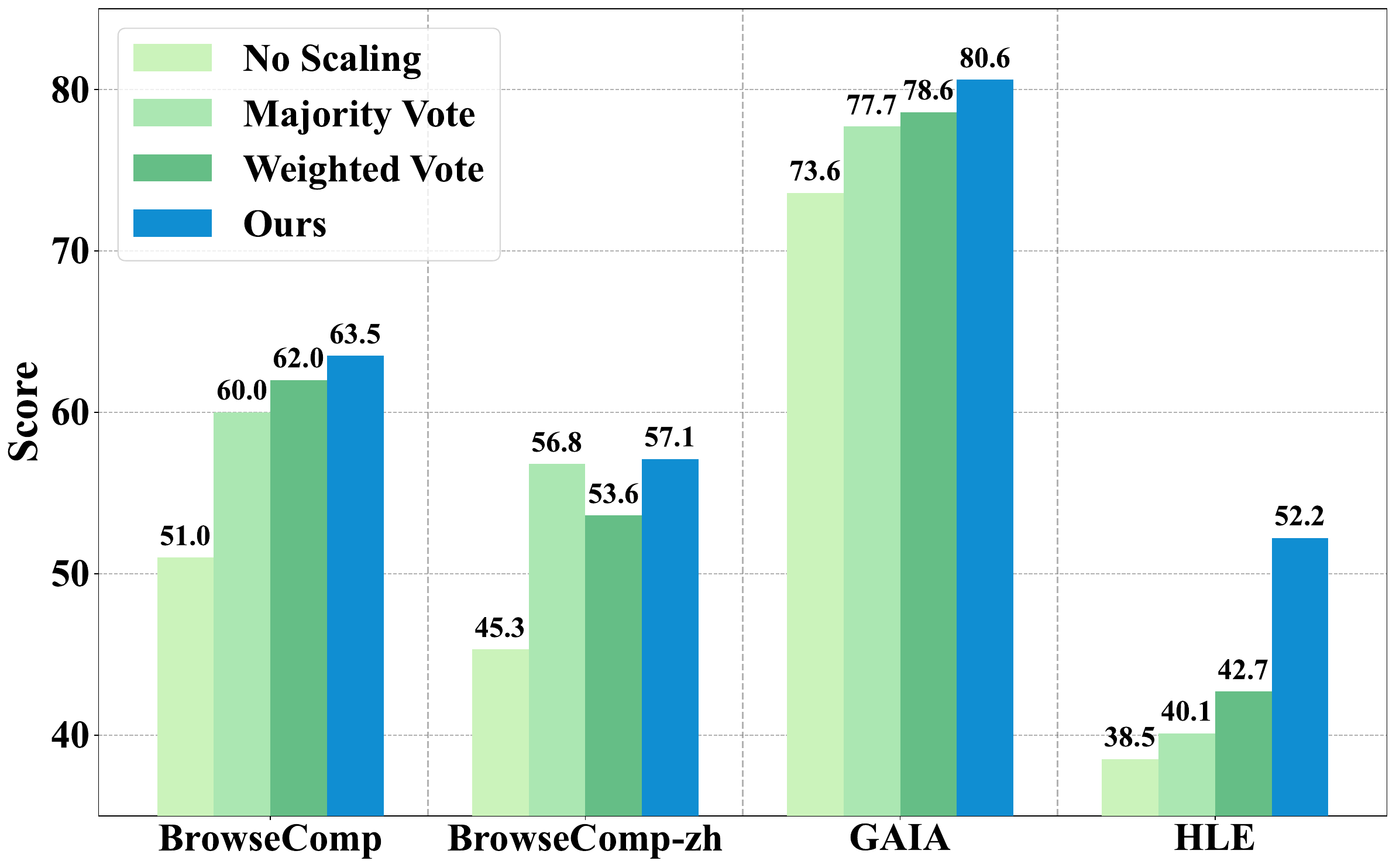}
    % \vspace{-2em}
    \caption{Performance gains from different answer generation methods, with sampling fixed to 8 from-scratch rollouts to isolate sampling (exploration) effects.}
    % \vspace{-2em} % 下方间距
    \label{fig:ablation_stage_2}
\end{wrapfigure}

In Section \ref{sec:function_region}, we examined the performance gains arising from the first-stage partial rollout of the proposed \textsc{ParallelMuse}. In this section, we isolate and analyze the effectiveness of its second-stage answer aggregation method, independently assessing its contribution to overall performance.

As shown in Figure \ref{fig:ablation_stage_2}, even without the sampling (exploration) gains from the first-stage partial rollout, the proposed \textit{Compressed Reasoning Aggregation} (the second-stage of \textsc{ParallelMuse}) alone yields the most significant improvement. Notably, this approach performs near-lossless compression over each agentic reasoning trajectory to efficiently integrate reasoning information without invoking additional tool calls for secondary verification. By maximizing the exploitation of existing sampled information during aggregation, it achieves a balanced improvement in both efficiency and solution quality.

\subsection{Efficiency Gains through Context Reuse and Trajectory Compression}

\begin{figure}[h]
    \centering
    \includegraphics[width=0.99\linewidth]{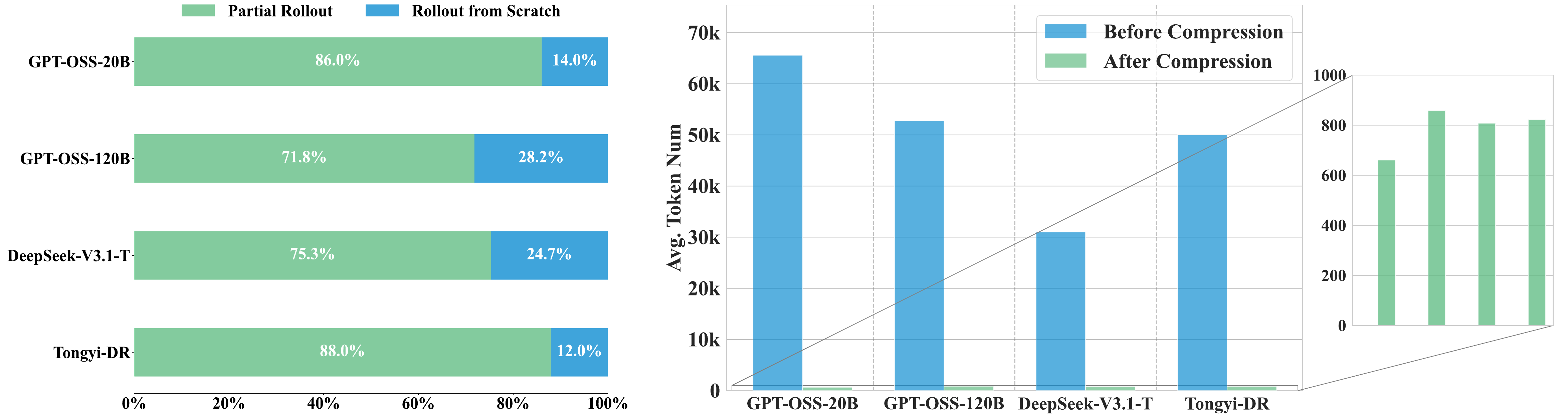}
    % \vspace{-2em}
    \caption{Efficiency gains using \textsc{ParallelMuse}. 
(i) (\textit{Left}) Token reduction through context reuse in our partial rollout method. 
We take the token consumption per trajectory of the from-scratch rollout as the baseline. 
The green bars represent the token cost after applying partial rollout (the numbers above indicate the ratio relative to the baseline), while the remaining blue bars show the proportion of tokens saved. 
(ii) (\textit{Right}) Comparison of context token usage before and after trajectory compression.}
    % \vspace{-2em} % 下方间距
    \label{fig:efficiency}
\end{figure}

We conduct a detailed analysis of the efficiency gains achieved by our proposed \textsc{ParallelMuse}, which primarily arise from two complementary sources:

\begin{enumerate}[label=\textbf{(\roman*)}]
    \item \textbf{Token reduction via context reuse in \textit{Functionality-Specified Partial Rollout}.}  
    As shown in Figure \ref{fig:efficiency} (\textit{Left}), our method (Partial Rollout) achieves up to \textbf{28\%} token savings by effectively reusing context instead of regenerating it from scratch (Rollout from Scratch).  
    The efficiency gain increases with sampling scale, indicating better scalability.

    \item \textbf{Context efficiency via trajectory compression in \textit{Compressed Reasoning Aggregation}.}  
    As shown in Figure~\ref{fig:efficiency} (\textit{Right}), compressing the agentic reasoning trajectory reduces context token usage by up to \textbf{99\%} relative to the full trajectory, achieving an almost complete compression. 
    This enables multi-trajectory reasoning aggregation within context limits and improves processing efficiency.
\end{enumerate}

% In summary, the proposed \textsc{ParallelMuse} is not a conventional test-time scaling strategy that improves performance by aggressively sacrificing efficiency. 
% By leveraging a task-informed design, it enhances overall reasoning effectiveness while eliminating most redundant or avoidable computation.
In summary, the proposed \textsc{ParallelMuse} is not a conventional test-time scaling strategy that improves performance by aggressively sacrificing efficiency. 
Instead, by leveraging a task-informed design, it scales computation where it matters most, allocating additional tokens and reasoning capacity as needed to high-utility regions while eliminating most redundant or avoidable computation.

\subsection{Impact of Model Capability on Compressed Reasoning Aggregation}

\input{table/02_stronger_model}

In the proposed \textsc{ParallelMuse}, the compression process in \textit{Compressed Reasoning Aggregation} can be viewed as extracting and reconstructing the agent’s internal information state graph $\mathcal{G}$ (defined in \eqref{graph}) from the full agentic reasoning trajectory.  This graph, described by the compressed report, encapsulates all information necessary for answer derivation. Hence, the quality of compression depends on the fidelity of this extraction and reconstruction, which directly affects subsequent aggregation performance.  

To examine whether a stronger model can perform higher-quality compression and yield better aggregation, we evaluate the setting where the \textit{Compressed Reasoning Aggregation} stage is executed by models stronger than those used for the first-stage partial rollout.  As shown in Table \ref{tab:stronger_model}, when the first-stage sampling is conducted with GPT-OSS-20B, replacing it with a stronger GPT-OSS-120B for the aggregation stage leads to a clear performance improvement. Further substitution with GPT-5 brings continuous gains, and a similar trend is observed on the Tongyi-DR-30B-A3B model, confirming that the compressed report effectively represents the agent’s internal information state graph and that higher-quality graph reconstruction enhances overall performance.
This result also suggests a practical insight for multi-agent design~\citep{han2024llm,li2024survey}: combining models of different strengths can balance efficiency and performance.

%% file: table/00_main_results.tex
\begin{table}[h]
\small
\centering
\caption{Overall performance. Scores marked with $\ddag$ represent full-benchmark results, whereas unmarked scores correspond to our benchmark settings. Both \textsc{ParallelMuse} and other parallel thinking baselines are evaluated under the default configurations as described in Section \ref{sec:setup}. The specific strategy for selecting functional regions in \textsc{ParallelMuse}’s partial rollout is discussed in Section \ref{sec:function_region}.}
\label{tab:main_result}
\resizebox{\columnwidth}{!}{
\setlength{\tabcolsep}{6pt} % 调整列间距
\renewcommand{\arraystretch}{1.3} % 调整行高
\begin{tabular}{l|l|c|c|c|c}
\toprule
\textbf{Model / Framework} & \textbf{Method} & \textbf{BrowseComp} & \textbf{BrowseComp-zh} & \textbf{GAIA} & \textbf{HLE} \\
\midrule
\multicolumn{6}{c}{\cellcolor{blue!20}\textit{\textbf{Closed-Source Deep Information-Seeking Agents}}} \\
\midrule
Claude-4-Sonnet & No Scaling & 12.2$^\ddag$ & 29.1 & 68.3 & 20.3$^\ddag$ \\
OpenAI-o3 & No Scaling & 49.7$^\ddag$ & 58.1 & 70.5 & 26.6$^\ddag$ \\
Kimi Researcher & No Scaling & -- & -- & -- & 26.9$^\ddag$ \\
OpenAI DeepResearch & No Scaling & 51.5$^\ddag$ & 42.9 & 67.4 & 26.6$^\ddag$ \\
ChatGPT Agent & No Scaling & 68.9$^\ddag$ & -- & -- & 41.6$^\ddag$ \\
\midrule
\multicolumn{6}{c}{\cellcolor{blue!30}\textit{\textbf{Open-Source Deep Information-Seeking Agents}}} \\
\midrule
\multirow{5.1}{*}{GPT-OSS-20B} & No Scaling & 30.9 & 28.6 & 63.4 & 24.2 \\
& Majority Vote & 44.0 & 38.8 & 69.9 & 24.2 \\
& Max \#Tool Call & 17.0 & 19.0 & 58.3 & 26.1 \\
& Weighted Vote & 41.0 & 37.0 & 68.9 & 31.2 \\
& \cellcolor{blue!10}\textbf{\textsc{ParallelMuse}} & \cellcolor{blue!10}\textbf{49.0} & \cellcolor{blue!10}\textbf{44.3} & \cellcolor{blue!10}\textbf{72.8} & \cellcolor{blue!10}\textbf{32.5} \\
\midrule
\multirow{5.1}{*}{GPT-OSS-120B} & No Scaling & 34.9 | 33.8$^\ddag$ & 36.0 & 74.3 & 36.3 \\
& Majority Vote & 48.5 & 46.7 & 77.7 & 43.3 \\
& Max \#Tool Call & 17.5 & 26.3 & 68.9 & 36.9 \\
& Weighted Vote & 48.0 & 45.7 & 82.5 & 45.2 \\
& \cellcolor{blue!10}\textbf{\textsc{ParallelMuse}} & \cellcolor{blue!10}\textbf{56.5} & \cellcolor{blue!10}\textbf{54.3} & \cellcolor{blue!10}\textbf{85.4} & \cellcolor{blue!10}\textbf{45.9} \\
\midrule
\multirow{5.1}{*}{DeepSeek-V3.1-T} & No Scaling & 23.2 & 36.1 & 61.0 & 25.0 | 21.7$^\ddag$ \\
& Majority Vote & 30.0 & 45.0 & 70.9 & 26.1 \\
& Max \#Tool Call & 17.5 & 28.0 & 57.3 & 27.4 \\
& Weighted Vote & 29.5 & 45.0 & 70.9 & 28.0 \\
& \cellcolor{blue!10}\textbf{\textsc{ParallelMuse}} & \cellcolor{blue!10}\textbf{39.0} & \cellcolor{blue!10}\textbf{50.2} & \cellcolor{blue!10}\textbf{74.8} & \cellcolor{blue!10}\textbf{37.6} \\
\midrule
\multirow{5.1}{*}{Tongyi-DR-30B-A3B} & No Scaling & 51.0 | 43.4$^\ddag$ & 45.3 & 73.6 & 38.5 | 32.9$^\ddag$ \\
& Majority Vote & 60.0 & 56.8 & 77.7 & 40.1 \\
& Max \#Tool Call & 41.0 & 36.3 & 75.7 & 38.2 \\
& Weighted Vote & 62.0 & 53.6 & 78.6 & 42.7 \\
& \cellcolor{blue!10}\textbf{\textsc{ParallelMuse}} & \cellcolor{blue!10}\textbf{65.0} & \cellcolor{blue!10}\textbf{57.1} & \cellcolor{blue!10}\textbf{79.6} & \cellcolor{blue!10}\textbf{52.2} \\
\bottomrule
\end{tabular}
}
\end{table}

%% file: table/01_partial_rollout.tex
\begin{table}[h]
\small
\centering
\caption{Performance comparison between full from-scratch rollouts and partial rollouts guided by functional-region uncertainty. Detailed configurations are listed in Table \ref{tab:hyperparams}.}
\begin{adjustbox}{width=1.0\linewidth}
\centering
\label{tab:partial_rollout}
\renewcommand\arraystretch{1.2} % row space
\begin{tabular}{l|l|c|c|c|c}
\toprule
\textbf{Agent Model} & \textbf{Functional Region} & \textbf{BrowseComp} & \textbf{BrowseComp-zh} & \textbf{GAIA} & \textbf{HLE} \\
\midrule
\multirow{4.3}{*}{GPT-OSS-120B} & From Scratch: No Region & 34.9 & 36.0 & 74.2 & 36.3 \\
\cmidrule(l){2-6}
% & No Partition & 35.5 & 41.4 & 74.8 & \jialong{TBD} \\
& Partial: Reasoning & 37.9 & \cellcolor{blue!10}\textbf{43.1} & 76.1 & 36.3 \\
& Partial: Exploration & \cellcolor{blue!10}\textbf{39.9} & 41.6 & \cellcolor{blue!10}\textbf{77.9} & \cellcolor{blue!10}\textbf{37.5} \\
& Partial: Mixed & 38.1 & 42.9 & 76.7 & 37.1 \\
\midrule
\multirow{4.3}{*}{DeepSeek-V3.1-T} & From Scratch: No Region & 23.2 & 36.1 & \textbf{61.0} & 25.0 \\
\cmidrule(l){2-6}
& Partial: Reasoning & \cellcolor{blue!10}\textbf{26.5} & \cellcolor{blue!10}\textbf{39.8} & 60.2 & \cellcolor{blue!10}\textbf{26.4} \\
& Partial: Exploration & 23.8 & 37.9 & \cellcolor{blue!10}60.6 & 25.3 \\
& Partial: Mixed & 23.4 & 38.1 & 60.3 & 25.1 \\
\bottomrule
\end{tabular}
\end{adjustbox}
\end{table}

%% file: table/02_stronger_model.tex
\begin{wraptable}{r}{8cm}
\small
\centering
\caption{Performance gains from using stronger models for \textit{Compressed Reasoning Aggregation}. Rollout configuration detailed in Table \ref{tab:hyperparams}.}
\begin{adjustbox}{width=1.0\linewidth}
\centering
\label{tab:stronger_model}
\renewcommand\arraystretch{1.2} % row space
\begin{tabular}{l|l|c}
\toprule
\textbf{Rollout Model} & \textbf{Aggregation Model} & \textbf{BrowseComp} \\
\midrule
\multirow{3.1}{*}{GPT-OSS-20B} & GPT-OSS-20B & 49.0 \\
& GPT-OSS-120B~\textcolor{darkgreen}{$\uparrow$} & 50.5 \\
& \cellcolor{blue!10}\textbf{GPT-5}~\textcolor{darkgreen}{$\uparrow$}\textcolor{darkgreen}{$\uparrow$} & \cellcolor{blue!10}\textbf{55.5} \\
\midrule
\multirow{2.1}{*}{Tongyi-DR-30B-A3B} & Tongyi-DR-30B-A3B & 65.0 \\
& \cellcolor{blue!10}\textbf{GPT-5}~\textcolor{darkgreen}{$\uparrow$} & \cellcolor{blue!10}\textbf{66.0} \\
\bottomrule
\end{tabular}
\end{adjustbox}
\end{wraptable}

%% file: sections/5_related_work.tex
\section{Related Work}

\subsection{Deep Information-Seeking Agents}

Deep information-seeking (IS) agents are autonomous systems designed to engage in multi-step interaction with external information environments (such as the web) and integrate retrieved data through reasoning in order to address complex knowledge-intensive tasks. 

The development of such agents has benefited from both proprietary and open-source initiatives. 
On the proprietary side, systems from major research organizations have set benchmarks for deep exploration and reasoning capabilities~\citep{openaidr,kimi_researcher,skywork2025deepresearch,perplexity,claude4,gemini2.5}, though their architectures and training protocols remain closed. 
On the open-source front, community efforts have advanced transparency and reproducibility in deep IS agent design~\citep{evolvesearch,webwalker,wu2025webdancerautonomousinformationseeking,li2025websailornavigatingsuperhumanreasoning,li2025websailorv2bridgingchasmproprietary,tao2025webshaper,geng2025webwatcher,sun2025simpledeepsearcher,asearcher,liu2025webexplorer,lu2025deepdive,tongyidr}, driving continuous progress in this domain.

In this work, we further explore the unique characteristics of deep IS agents and propose \textsc{ParallelMuse} to exploit these properties more effectively, thereby enhancing both capability and efficiency.

\subsection{Parallel Thinking for Test-Time Scaling}

Parallel thinking~\citep{wang2025survey} serves as an effective test-time scaling strategy for reasoning, particularly in agentic settings that require deep and complex interactions. It generates multiple reasoning trajectories to capture diverse reasoning behaviors and jointly determines the final answer.

Conceptually, parallel thinking follows a two-stage paradigm~\citep{li2025raspberry}: \textit{exploratory sampling} and \textit{answer generation}. The first stage explores diverse reasoning paths through independent sampling~\citep{wei2022chain,zeng2025pushing}, structured rollouts, or intermediate partial rollouts~\citep{skywork2025deepresearch,arpo,hou2025treerl,li2025treepo} where branches share context but remain flexible. In agentic reasoning with vast exploration spaces, independent and partial rollouts are generally more effective and efficient than structured ones. The second stage focuses on synthesizing results via either answer selection~\citep{self_consistency,fu2025deep}, which is efficient but often biased, or answer aggregation~\citep{jiang2023llm,liang2024encouraging,zhang2025and,qiao2025webresearcher}, which is more stable but challenged by the need to identify which intermediate reasoning contributes most to the final outcome.

Most existing parallel thinking methods inherit the assumptions of pure reasoning tasks. Building on a detailed analysis of agentic reasoning, especially in deep IS tasks, we propose \textsc{ParallelMuse}, a paradigm that fully leverages these properties to more effectively unlock the potential of deep IS agents.

%% file: sections/6_conclusion.tex
\section{Conclusion}

This work investigates the challenges of applying parallel thinking to deep information-seeking (IS) agents. Conventional parallel thinking strategies often waste computation through redundant rollouts and struggle to integrate long-horizon reasoning due to limited context capacity. Building on an in-depth analysis of the characteristics of deep IS tasks, we incorporate these insights into method design and propose \textsc{ParallelMuse}, a two-stage paradigm that enhances both exploration efficiency and reasoning aggregation. Experimental results across multiple open-source agents and benchmarks demonstrate that \textsc{ParallelMuse} achieves substantial performance improvements while greatly reducing exploratory token consumption, highlighting its effectiveness for efficient and scalable deep IS reasoning.

%% file: sections/7_limitation.tex
\section{Limitations and Future Work}

In this work, we focus primarily on question-answering–oriented deep IS tasks, where the toolset is limited to Search and Visit. While this configuration is optimal for deep IS tasks, more general agentic tasks often involve a broader range of tools~\citep{fang2025towards}, leading to substantially larger exploration spaces. 
Designing effective parallel thinking strategies under such complex tool configurations to extend applicability to general agentic settings remains an open direction for future research.